\documentclass{bmvc2k}

\title{Cross-Domain Tracker Adaptation Without Target-Domain Labels via Vision-Language Agents}

\addauthor{Daniel Davila}{ddavila2@cisco.com}{1}
\addauthor{Ravikumar Balakrishnan}{ravikuba@cisco.com}{1}
\addauthor{Mike Cochran}{mikcochr@cisco.com}{1}

\addinstitution{
 Cisco Systems, Inc. \\
 San Jose, CA
}

\runninghead{D Davila, R Balakrishnan, M Chochran}{VLM-Based Tracker Adaptation}

\usepackage{booktabs}

\begin{document}

\maketitle

\begin{abstract}
We present a system that uses a Vision-Language Model (VLM) as a diagnostic agent for adapting a detect-to-track pipeline to a new target domain without access to target-domain labels. Rather than optimizing against annotated metrics, the VLM directly inspects rendered tracking outputs, identifies visual failure modes, and recommends parameter updates through an iterative tuning loop. We first demonstrate that ground-truth-supervised hyperparameter transfer can be brittle. On MOT17→MOT20, applying a source-derived oracle configuration reduces mean HOTA by 0.090, from a target-domain ceiling of 0.357, to 0.267. Without using any target-domain labels, our VLM-based tuner recovers 67.8\% of this lost headroom, finishing within 0.029 HOTA of the target ceiling; on the highest-density target sequence, it recovers up to 86.7\%. We further show that label-free Bayesian optimization with handcrafted proxy objectives struggles under large domain shifts and can degrade configurations that are already strong. In contrast, the VLM tuner acts selectively: when its visual diagnosis reveals no clear failure mode, it declines to modify the configuration, preserving performance on easy transfers while improving hard ones. Finally, we characterize the conditions under which this approach succeeds, namely, when domain shift manifests through exposed detection-level parameters, versus where it is less effective, such as MOT17→DanceTrack, where the source oracle is already near-optimal.
\end{abstract}

\section{Introduction}
\label{sec:intro}


Detect-to-track systems deployed in new visual domains typically suffer performance degradation due to domain shift in scene density, camera viewpoint, and target dynamics. In the wild, practitioners often tune trackers by visually inspecting the output and adjusting hyperparameters until results appear qualitatively acceptable. This is a time-consuming, error-prone process that cannot be easily automated or repeated when operating conditions change. Additionally, similar challenges arise when scene dynamics drift over time, though we focus on the cross-domain case here. While methods exist to detect out-of-domain deployment in online systems, few address how to adapt the hyperparameter space to the new domain without ground truth or human feedback.

\begin{figure}[t]
  \centering
  \includegraphics[width=\linewidth]{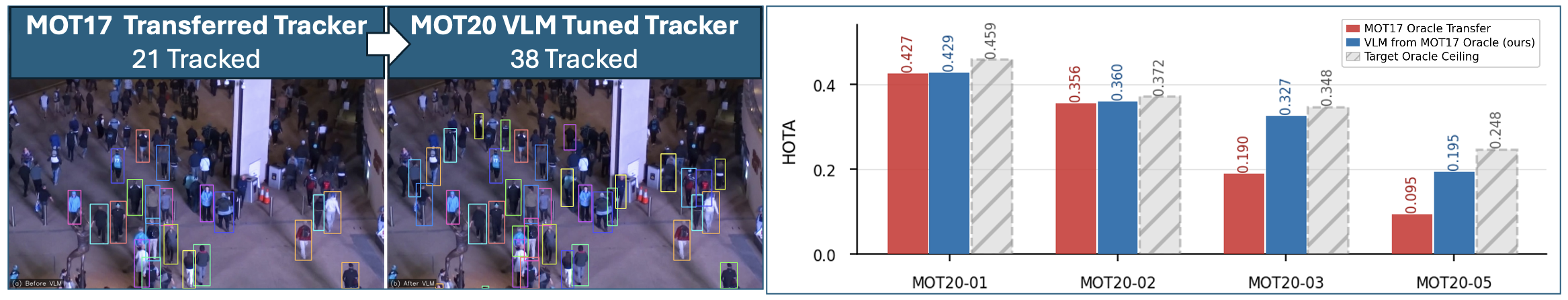}
  \caption{Cross-domain tracker adaptation without target-domain labels. (Left) A MOT20-03 frame under MOT17-tuned
parameters, and then again after VLM-guided adaptation
on the target sequence without target-domain labels. Tuning increases number of persons tracked in difficult scene. (Right) Per-sequence HOTA on MOT20: our
VLM-tuned configuration (blue) closes most of the gap
between oracle transfer (red) and the target per-sequence
oracle ceiling (gray hatched), with the largest gains on the
hardest sequences MOT20-03/05.}
\label{fig:teaser}
\end{figure}

In this paper, we first establish the brittleness of scene-specific tracker hyperparameters. For each scene, we compute an oracle configuration by performing a grid search over the tracker’s exposed parameter space, thereby defining an empirical performance ceiling for that scene. We do this using ground-truth annotations for videos from MOT17, MOT20, and DanceTrack. We then transfer each oracle configuration to other videos and measure the resulting degradation relative to the oracle configuration computed directly on the target scene. Our results show a consistent drop in HOTA and other key tracking metrics under cross-domain transfer, with the degradation most pronounced when moving to high-density scenes such as MOT20-03 and MOT20-05, or from lower- to higher-motion regimes, as in transfers from MOT to DanceTrack.

Based on these findings and the observation that subject matter experts often can and do perform manual qualitative adjustments to live tracking systems, we propose a tracker adaptation framework using vision-language feedback that, starting from
a source-domain oracle and using no target-domain labels,
automatically tunes the tracker on a new target domain. A representative result from this method is shown in Figure~\ref{fig:teaser}. The
VLM consumes purpose-built visual diagnostics including detection
collages, scene-level dot panels, and track-identity strips, which surface failure modes across detection (missed
detections, false positives) and tracking (identity switches,
fragmentation). Without access to target-domain ground truth,
the VLM prescribes targeted parameter adjustments through a
sequential two-phase protocol (detection → tracking),
recovering 67.8\% of the per-sequence oracle headroom on
MOT17$\rightarrow$MOT20 and reaching within 0.029 HOTA of
the target-domain oracle. 

The contributions of this paper are summarized as follows: 
\begin{itemize}
\item We demonstrate that ground-truth-optimized tracker configurations are brittle under domain shift, with supervised oracle transfer from MOT17 to MOT20 falling 0.090 HOTA below the target-domain per-sequence oracle, establishing that careful per-domain tuning does not transfer sufficiently.

\item We propose a tracker adaptation framework using vision-language feedback that, starting from a source-domain oracle and using no target-domain labels, recovers up to 86.7\% of per-sequence oracle headroom on the highest-density target sequence and 67.8\% mean recovery across MOT17→MOT20.

\item We release as open source the complete toolkit and a full prompt-engineering experimentation log (40+ iterations) documenting the design phenomenology of VLM-in-the-loop optimization, including failure modes, prompt evolution, and feedback representation design Available at https://github.com/dsdavila/vlm\_tracker\_tuner. 
\end{itemize}


\section{Related Work}
\label{sec:related}

\subsection{Multi-Object Tracking Pipelines}
\label{sec:rw-mot}

The dominant paradigm in multi-object tracking is tracking-by-detection. A
per-frame detector produces candidate boxes that a separate stage associates
into tracks. The SORT family \cite{bewley2016simple, wojke2017simple,
zhang2022bytetrack, aharon2022botsort} is a well-studied baseline, with ByteTrack
\cite{zhang2022bytetrack} and BoT-SORT \cite{aharon2022botsort} the current
standard on pedestrian benchmarks. These systems expose a small set of
operating parameters (confidence thresholds, NMS aggressiveness, IoU gates,
lost-track survival) and are known to be sensitive to them. Additionally, operating-point sensitivity to scene density is well documented across the
family. We adopt YOLOv11-Large \cite{ultralytics_yolo} with ByteTrack as a
representative, widely-used instance of this paradigm.

\subsection{Domain Adaptation for Tracking}
\label{sec:rw-da}

The standard approach to domain shift in MOT adapts the underlying
\emph{model} on the target domain. DARTH~\cite{segu2023darth} introduces a
test-time adaptation framework for MOT, jointly adapting detection (via a
self-supervised detection-consistency loss) and instance appearance (via a
patch contrastive loss). Path Consistency~\cite{lu2024path} learns object
matching without identity labels by enforcing consistency across
frame-skipping observation paths. GeneralTrack~\cite{qin2024generaltrack}
takes a generalization view, with an architecture that handles cross-scenario
motion and appearance variability, and PASTA~\cite{mancusi2024pasta} trains
per-attribute expert modules composed in parameter space for new domains. LTTrack~\cite{yu2023generalizing} is a trainable end-to-end tracker with a learned language
representation, trained on the source domain to generalize to the target. It uses weight updates based on a training signal. 

These approaches all bridge the domain gap by modifying the model. We instead
address \emph{operating-parameter} adaptation with detector and tracker
weights fixed and only runtime thresholds and gates adjusted,
the regime practitioners face when retraining is impractical.
SQE~\cite{huang2020sqeselfqualityevaluation} is closest in spirit, using
Bayesian optimization against a tracker-output proxy for label-free
cross-domain adaptation; we include a baseline of this kind
(Section~\ref{sec:results}) and find it underperforms VLM-guided adaptation
and can degrade already-good configurations. To our knowledge, no prior work
addresses operating-parameter adaptation under no target-domain labels with
visual feedback of tracker output.

\subsection{Hyperparameter Optimization and LLM Agents}
\label{sec:rw-hpo}

Classical hyperparameter optimization, such as Bayesian optimization and
Population-Based Training \cite{snoek2012bayesopt, jaderberg2017pbt}, treats
HPO as black-box search guided by a scalar reward computed against a labeled
validation set. That labeled set is load-bearing as it defines the optimization
target. Without it these methods have no direction to search in. A recent
line of work casts Large Language Models as HPO agents; AgentHPO
\cite{liu2025agenthpo} has an LLM propose configurations, observe performance,
and refine its proposals. This replaces the search heuristic with
language-model reasoning but leaves the feedback signal a supervised metric
computed against ground truth. The LLM makes search more sample-efficient
and interpretable, but does not remove the labeled-data requirement.

Our work differs from previous  works here. We assume no target-domain labels,
so the signal driving optimization is not a supervised metric but a structured
set of \emph{visual diagnostics} (detection collages, dot panels, NMS-orphan
overlays) consumed by a VLM in lieu of ground truth. This is what motivates a
VLM rather than an LLM as the diagnostic agent. The strongest available signal
about target-domain performance is what the tracker's output \emph{looks
like}, which we show in our ablations.

\subsection{VLMs as Diagnostic Agents}
\label{sec:rw-vlm}

A growing body of work uses Vision-Language Models as evaluators or critics
over visual content. Prometheus-Vision \cite{lee2024prometheusvision} trains a
VLM evaluator that scores generated content against user-defined criteria, and
Critic-V \cite{zhang2025criticv} trains critic VLMs to catch errors in other
VLMs' multimodal reasoning. In these works, the VLM acts as an evaluator on a
fixed system's output, producing an assessment, critique, or score. We take
the framing one step further: the VLM does not merely \emph{assess} output, it
prescribes \emph{interventions} on the system's operating parameters, returning
structured diagnoses together with parameter adjustments applied to the tracker
on the next iteration. To our knowledge, this is the first use of a VLM in a
closed-loop control role for a perception pipeline, rather than as an
open-loop evaluator.

\section{Method}
\label{sec:method}

\subsection{System Overview}
\label{sec:overview}

\begin{figure}[t]
  \centering
  \includegraphics[width=\linewidth]{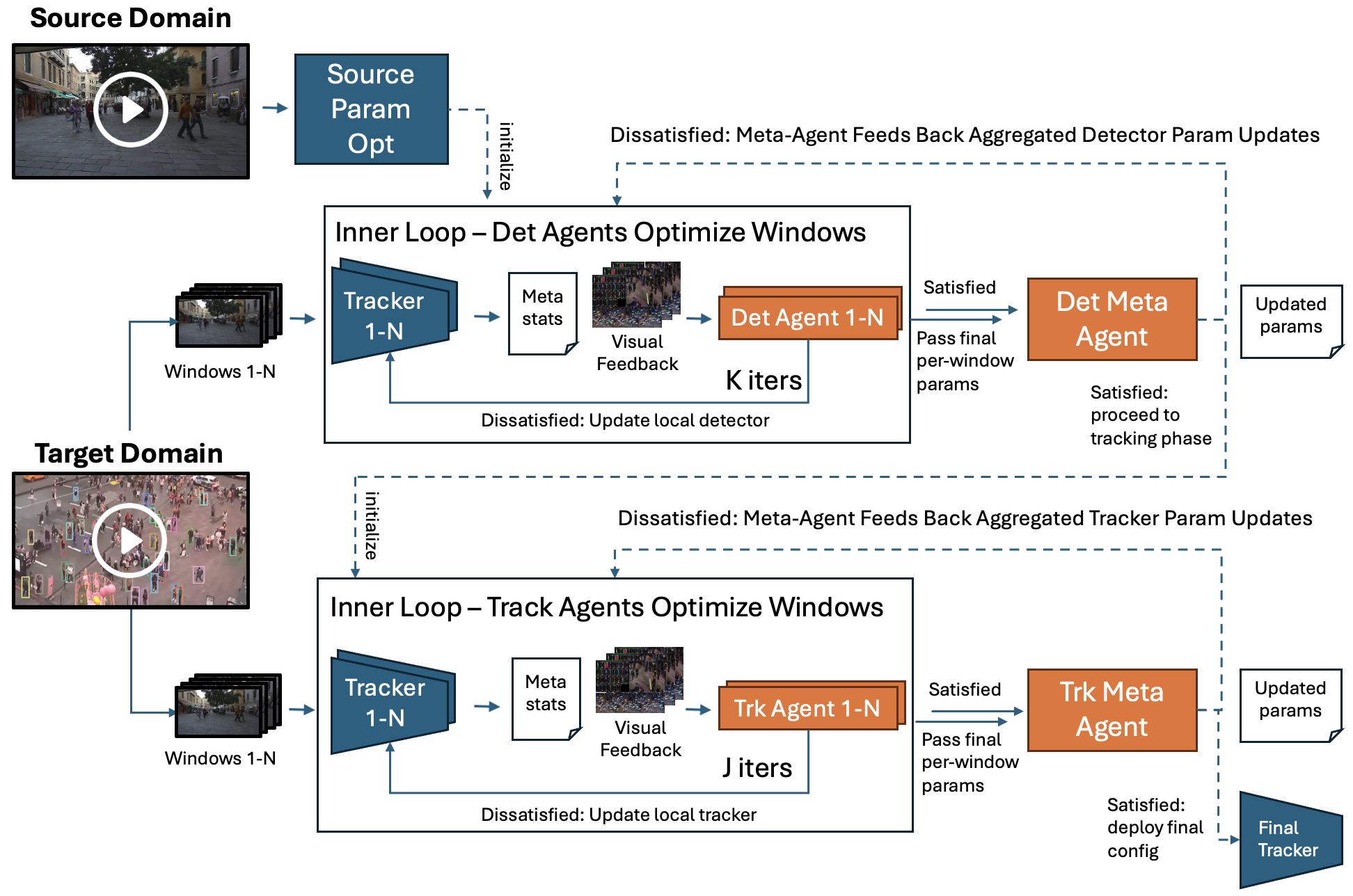}
  \caption{Overview of VLM Tracker Tuner for cross-domain tracker
  adaptation without target-domain ground truth.}
  \label{fig:arch}
\end{figure}

The full system architecture is illustrated in Figure~\ref{fig:arch}. Our
method takes a fixed detect-to-track pipeline and a new target domain, and
produces a target-adapted parameter configuration without using any
target-domain labels, in four stages. Importantly, tuning runs asynchronously with respect to the tracker loop.

\paragraph{Stage 1 --- Source-domain initialization.}
We first use the labeled \emph{source} domain to compute a supervised
initialization for the tracker parameters. Specifically, for each source
sequence, we obtain an \emph{oracle} configuration by grid search over the
tracker's exposed parameter space, as described in Section~\ref{sec:oracle}.
This configuration serves as the source-supervised reference point from which
target-domain adaptation begins. After this initialization step, the method
does not use source or target labels.

\paragraph{Stage 2 --- Two-phase target-domain adaptation.}
Adaptation on the unlabeled \emph{target} domain is performed in two sequential
phases: the \emph{detection phase}, which updates detection-related parameters,
and the \emph{tracking phase}, which updates tracking and association
parameters. This decomposition reduces the complexity of each VLM reasoning
step and separates detection failures from downstream association failures.
Each phase uses a phase-specific visual diagnostic suite and failure-mode
taxonomy, but both phases follow the same adaptation structure.

\paragraph{Stage 3 --- Window-level diagnosis and sequence-level aggregation.}
Each adaptation phase consists of an inner window-level diagnosis step and an
outer sequence-level aggregation step. The target sequence is partitioned into
$N$ temporal windows. In the inner step, each window is analyzed independently
by a VLM-based diagnostic agent, which observes structured visual diagnostics
rendered from the current tracker output, applies a phase-specific diagnostic
procedure, and returns both a failure-mode diagnosis and a parameter
recommendation. In the outer step, a meta-agent aggregates the $N$ window-level
recommendations into a single sequence-level parameter update. The updated
configuration is then applied to all windows in the next iteration.

\paragraph{Stage 4 --- Termination and deployment.}
Each phase terminates when all window-level agents return one of two terminal diagnosis. The first, \texttt{clean}, indicates that no visible failure remains. The second, \texttt{no\_lever}, indicates that the observed failure is not addressable by
the exposed parameters. A phase also terminates if its phase-specific iteration
budget is exhausted. After the detection phase terminates, its final
configuration initializes the tracking phase. The final configuration after the
tracking phase is deployed as the target-adapted tracker.

The remainder of this section zooms in on each component: the pipeline and
parameter space (\ref{sec:pipeline}), the source oracle (\ref{sec:oracle}),
the adaptation loop mechanics (\ref{sec:loop}), and the two phase-specific
diagnostic designs (\ref{sec:detection-phase},~\ref{sec:tracking-phase}). The
experimental setup and Bayesian-optimization baseline appear in
Section~\ref{sec:experimental-design}, with results in
Section~\ref{sec:results}.

\subsection{Pipeline and Tunable Parameter Space}
\label{sec:pipeline}

We use a standard detect-to-track pipeline: a YOLOv11-Large detector followed
by ByteTrack association. The detector is used without architectural
modification and restricted to the person class. The pipeline exposes eight
tunable parameters, partitioned according to the two adaptation phases. The
\textit{detection-phase} parameters are \texttt{conf\_thresh}, the confidence
threshold below which YOLO outputs are discarded, and \texttt{nms\_thresh}, the
IoU threshold above which Non-Maximum Suppression removes the lower-confidence
detection. The \textit{tracking-phase} parameters ---
\texttt{match\_thresh}, \texttt{spawn\_thresh}, \texttt{max\_lost\_age},
\texttt{velocity\_decay}, \texttt{min\_hits}, and
\texttt{appearance\_thresh} --- control detection-to-track association, track
initialization, persistence through missed detections, motion extrapolation,
track-emission gating, and appearance-based re-identification, respectively.

\subsection{Source-Domain Oracle Computation}
\label{sec:oracle}

For each source sequence, we compute a per-sequence oracle by grid search over
the parameter space of \ref{sec:pipeline}, evaluating each grid point against
ground-truth-supervised HOTA and selecting the HOTA-maximizing configuration.
The grid resolution is chosen to bound search cost while sampling each
parameter densely enough that the oracle approaches the local performance
ceiling within the exposed parameter space. The oracle serves as both the
GT-supervised transfer baseline and the initialization point for VLM-guided
adaptation.

\subsection{Adaptation Loop Architecture}
\label{sec:loop}

Both the per-window agents and the meta-agent use the same VLM,
Gemma-4-31B, served through a local inference deployment; their distinct roles
are specified entirely by the phase-specific system prompt and diagnostic
suite with no shared context between agents. Each window agent also receives a small set of lightweight calibration
metrics computed only from tracker outputs. The prompt instructs the agent to use these metrics only to scale the
magnitude of parameter changes, never to set their direction. The meta-agent
observes all per-window proposals together with the visual evidence motivating
them, and resolves conflicts, e.g., one window proposing to lower
\texttt{conf\_thresh} while another proposes to raise it, by reasoning over
the joint evidence rather than averaging recommendations.

A phase terminates when all per-window agents return \texttt{clean} or
\texttt{no\_lever} in the same iteration, or when the phase-specific iteration
budget is exhausted. \texttt{clean} indicates that no visible actionable
failure remains in the window; \texttt{no\_lever} indicates that the observed
failure cannot be addressed by the exposed parameters. The detection-phase
output configuration initializes the tracking phase. The complete system
prompts, prompt-engineering log with more than 40 iterations, panel rendering
toolkit, and representative transcripts for successful and failed runs are
provided as supplementary material.

\subsection{Detection Phase: Design}
\label{sec:detection-phase}

\begin{figure}[t]
  \centering
  \includegraphics[width=\linewidth]{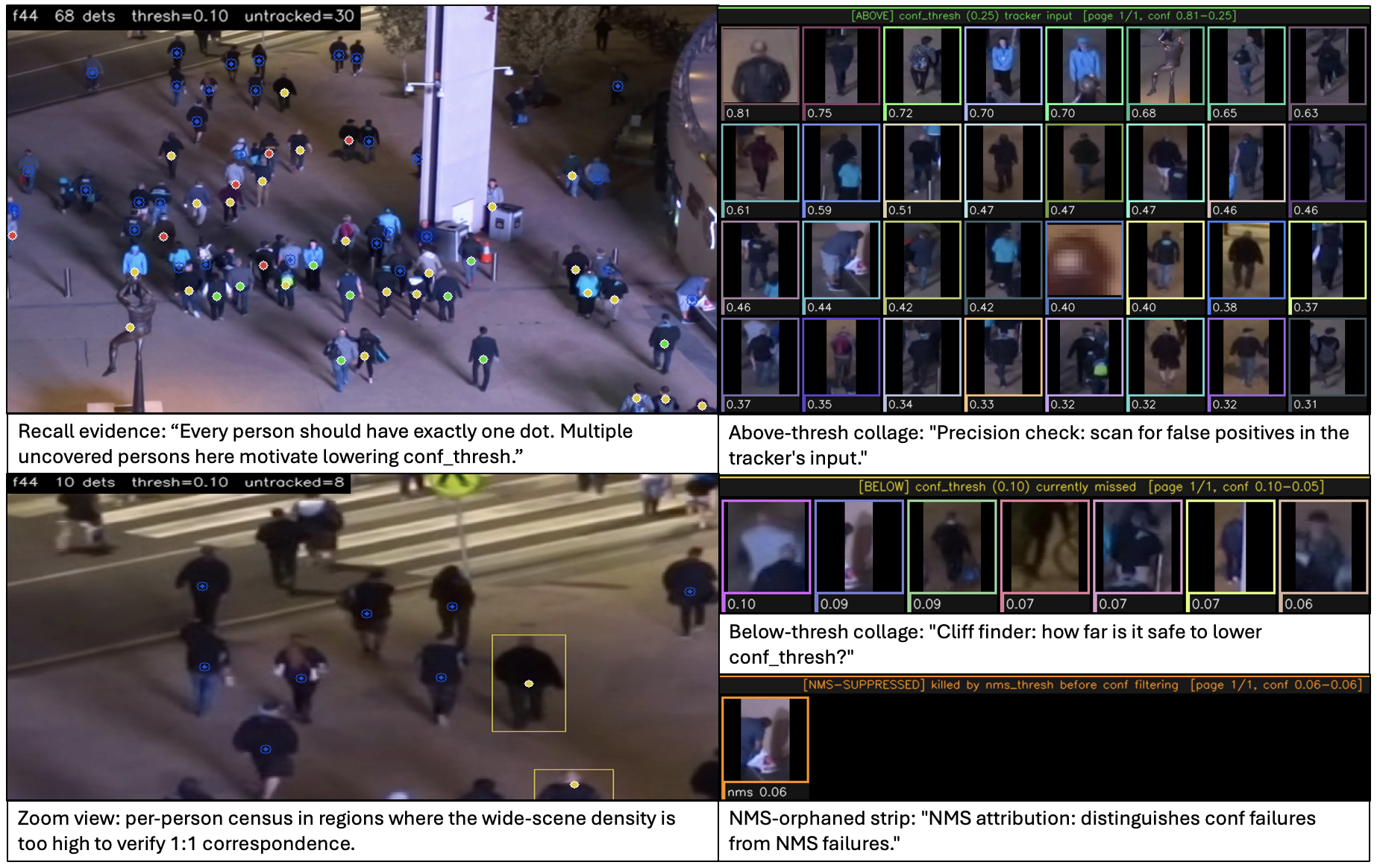}
  \caption{Detection-phase visual diagnostic suite. Each window of
  tracker output is rendered into five panel types delivered to the VLM as a
  single multimodal prompt: a full-scene dot panel and a zoomed dot panel
  (left) for recall evidence, above- and below-threshold collages (right) for
  precision and quality-cliff evidence, and an NMS-orphaned strip (bottom
  right) attributing suppressed detections to \texttt{nms\_thresh}.}
  \label{fig:detection_suite}
\end{figure}

The detection-phase agent applies a four-step diagnostic to each window, with
each step grounded in a specific panel of Figure~\ref{fig:detection_suite}.
\textbf{(1) Recall census:} the agent first enumerates persons in the
full-scene and zoomed \textit{dot panels}, which overlay confidence-encoded
detections on raw frames. This precedes inspection of box-level detector
outputs and reduces anchoring on existing boxes. \textbf{(2) Precision check:}
it scans the \textit{above-threshold collage}, containing all detections
currently passed to the tracker and sorted by confidence, for false positives.
\textbf{(3) Cliff search:} if Step~1 identifies a missed person not explained
by existing detections, the agent inspects the \textit{below-threshold collage}
(sub-threshold crops sorted closest-to-threshold first) to estimate how far
\texttt{conf\_thresh} can be safely lowered; this gate prevents duplicate
detections from spuriously lowering the threshold. \textbf{(4) Diagnosis under
hard gates:} using the \textit{NMS-orphaned strip}, which isolates detections
suppressed by NMS before confidence filtering, the agent selects the appropriate
lever, \texttt{conf\_thresh} or \texttt{nms\_thresh}, and attributes recall
failures to the corresponding cause.

Alongside the panels, the agent receives \texttt{near\_threshold\_fraction} (detections within $0.10$ of
\texttt{conf\_thresh}), \texttt{conf\_p50\_margin} (median-confidence distance
from threshold), and active-track population statistics. These gate only the
\emph{magnitude} of parameter changes, never their direction, which is set by
the panels. The relative contributions of visual and scalar evidence, and of
the two phases, are ablated in Section~\ref{sec:results}; the full agent
prompts appear in the supplementary. A representative trace is shown in
Figure~\ref{fig:trace_example}, with additional traces and failure cases in
the supplementary.

\begin{figure}[t]
  \centering
  \fbox{\parbox{0.95\columnwidth}{
    \small
    \textbf{Round:} 0 \quad\textbf{Window:} MOT20-03 frame 44

    \medskip
    \textbf{Initial config:} \texttt{conf\_thresh=0.25},
    \texttt{nms\_thresh=0.45}

    \medskip
    \textbf{Diagnosis:} \texttt{false\_negative\_detections}
    \quad (confidence 0.9)

    \medskip
    \textbf{Recommendation:} raise \texttt{nms\_thresh}:
    $0.45 \rightarrow 0.65$

    \medskip
    \textbf{Reasoning (verbatim):} \textit{``The dot panels
    and zoomed crops reveal significant under-detection (FNs)
    in a crowded scene. Crucially, multiple `nms\_orphan'
    dashed orange boxes are present [in the scene panels].
    These orphans represent people the detector found with
    high confidence but were killed by NMS because they
    overlapped with another detection. Because these targets
    were suppressed by NMS \textbf{before} the confidence
    filter, lowering \texttt{conf\_thresh} cannot recover
    them. To fix this, \texttt{nms\_thresh} must be raised
    to be less aggressive in crowded environments. While
    there is some room to lower \texttt{conf\_thresh} based
    on the quality cliff ($\sim$0.12), the primary driver of
    misses here is NMS suppression in the crowd.''}
  }}
  \caption{Representative VLM diagnostic trace (MOT20-03, frame 44, round 0).
  The agent identifies NMS-orphan markers, attributes the recall failure to
  NMS rather than \texttt{conf\_thresh}, and raises \texttt{nms\_thresh}
  accordingly. Image-index references in the original output are replaced with
  bracketed descriptors for clarity.}
  \label{fig:trace_example}
\end{figure}

\subsection{Tracking Phase: Design}
\label{sec:tracking-phase}

\begin{figure}[t]
  \centering
  \includegraphics[width=\linewidth]{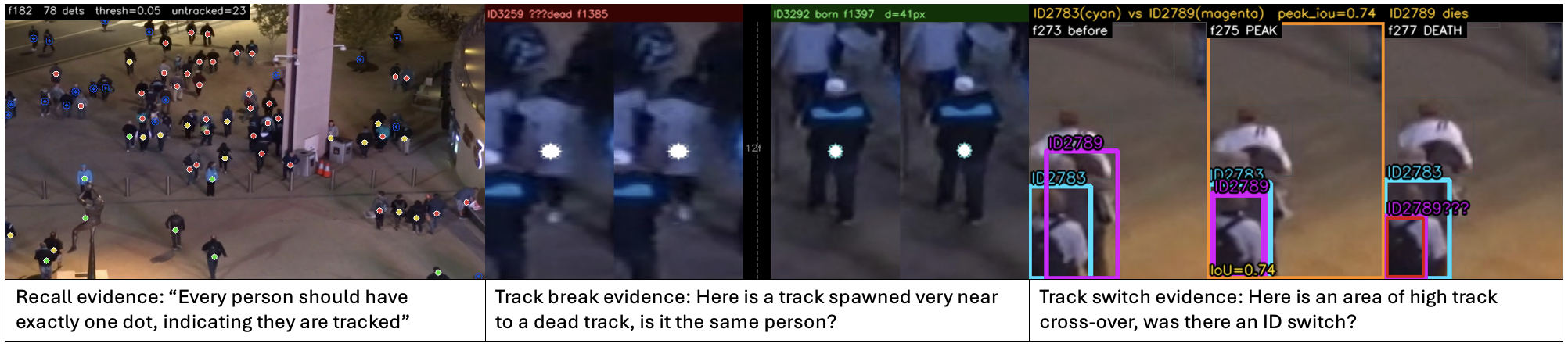}
  \caption{Tracking-phase visual diagnostic suite. Three panel types
  per window: (left) the scene/dot panel for recall evidence; (middle) track-break
  visualizations showing spawns near a dead-track endpoint to test
  whether a track was broken and re-established; and (right) track-approach
  visualizations showing likely crossover points, to check for ID
  switching.}
  \label{fig:tracking_suite}
\end{figure}

The tracking phase follows the same per-window-plus-meta-agent template, with
three substitutions: the parameter set is \{\texttt{match\_thresh},
\texttt{spawn\_thresh}, \texttt{max\_lost\_age}, \texttt{velocity\_decay},
\texttt{min\_hits}, \texttt{appearance\_thresh}\}, the diagnostics target
track-level rather than detection-level failures, and the taxonomy covers
identity and continuity failures. As shown in Figure~\ref{fig:tracking_suite}, the agent receives an enhanced
dot panel, re-ID death$\rightarrow$birth panels, and closest-approach strips.
The dot panel uses green/yellow/red filled circles for active tracks and blue
hollow circles for untracked detections. The re-ID panels pair a dying track's
final crops with a nearby new track's first crops to distinguish fragmentation
from coincidental proximity. The closest-approach strips show a track pair at
peak IoU across three frames to expose ID switches at crossings.
Table~\ref{tab:tracking-mapping} summarizes the failure-mode-to-parameter
decision tree; each mode maps to one or two primary levers, with secondary
fallbacks when the primary lever is exhausted across prior iterations.

\begin{table}[t]
\centering
\footnotesize
\setlength{\tabcolsep}{6pt}
\renewcommand{\arraystretch}{1.12}
\begin{tabular}{p{0.38\linewidth}p{0.50\linewidth}}
\hline
\textbf{Failure mode} & \textbf{Primary parameter response} \\
\hline
\texttt{id\_switch}
& $\downarrow$ \texttt{match\_thresh} \\
\texttt{ghost\_tracks}
& $\uparrow$ \texttt{min\_hits}, $\uparrow$ \texttt{spawn\_thresh} \\
\texttt{fragmented\_tracks}
& $\uparrow$ \texttt{max\_lost\_age}; then $\downarrow$ \texttt{velocity\_decay} \\
\texttt{frequent\_id\_switches}
& $\uparrow$ \texttt{match\_thresh}, $\downarrow$ \texttt{max\_lost\_age} \\
\texttt{track\_drift}
& adjust \texttt{velocity\_decay} \\
\hline
\end{tabular}
\caption{Tracking-phase failure-mode-to-parameter mapping. Arrows indicate
direction; magnitude is sized by lifecycle statistics
(death$\rightarrow$birth pairs, conf-at-death sequences). Secondary levers and
fallback rules omitted; see released prompts.}
\label{tab:tracking-mapping}
\end{table}

\section{Experiments}

\subsection{VLM Domain Transfer}
\label{sec:experimental-design}

We evaluate cross-domain transfer across the three-way
source/target matrix formed by MOT17, MOT20, and
DanceTrack ($3 \times 3 = 9$ source/target pairs,
including same-domain self-transfer). For each pair we
report HOTA under three conditions: \textit{oracle
transfer} (apply the source per-sequence oracle to the
target without further tuning; requires source-domain
GT only), \textit{VLM from oracle} (initialize from the
source oracle, then run the full two-phase VLM
adaptation loop on the target without target-domain
labels), and the \textit{target per-sequence oracle}
(grid search on the target sequence directly; requires
target-domain GT and serves as the
within-parameter-space ceiling). We additionally report
agent ablations isolating the detection and tracking
phases to attribute gains across the two-phase
protocol, as well as the contributions of the visual and text feedback mechanisms. We evaluate on the MOT17 training set (7
FRCNN sequences), the MOT20 training set (4
sequences), and the DanceTrack validation set (8
sequences), using TrackEval~\cite{trackeval} version
1.1.0 with HOTA as the primary metric.

The detect-to-track pipeline uses YOLOv11-Large
\cite{ultralytics_yolo} with pretrained
COCO weights at $1280 \times 736$ input resolution,
followed by the BoxMOT implementation of ByteTrack
\cite{zhang2022bytetrack}. The VLM is Gemma 4 31B
\cite{google2026gemma4} (instruction-tuned variant
\texttt{google/gemma-4-31B-it}), a 31-billion-parameter
dense multimodal model released by Google DeepMind in
April 2026, accessed through an OpenAI-compatible
local inference endpoint with max\_tokens=2048,
vision input enabled, and the endpoint's default
temperature of 1.0. Each adaptation run uses 3 detection-phase
windows and 3 tracking-phase windows selected by
event-targeted sampling rather than uniform
partitioning, with a base window length of 300 frames
and an adaptive floor of 143 frames for short
sequences. The per-phase revisit budget is
$K_{\text{det}} = 3$ and $K_{\text{trk}} = 4$ revisits
per window; only the meta-agent's decision writes the
next iteration's tracker configuration, ensuring
per-window agents propose against a frozen baseline
rather than overwriting each other. Inference was conducted
on an NVIDIA L40S GPU; VLM calls average 22.9~s per
call and the end-to-end adaptation pipeline averages
10.1~s per window across cached and fresh inferences.

\subsection{Comparison to Bayesian Optimizer}
\label{sec:bo-exp}
As a label-free optimizer baseline, we run Bayesian optimization over the same
exposed parameter space as the oracle grid search
(Section~\ref{sec:oracle}), but replace the ground-truth HOTA objective with a
scalar proxy computed entirely from tracker output:
\begin{equation}
\label{eq:bo-proxy}
\mathrm{proxy} \;=\; \frac{n_{\mathrm{tracks}} \cdot \overline{\ell}}
{1 + \rho_{\mathrm{frag}}},
\end{equation}
where $n_{\mathrm{tracks}}$ is the number of confirmed tracks,
$\overline{\ell}$ is their mean lifetime in frames, and $\rho_{\mathrm{frag}}$ is the fragmentation ratio, defined as the
number of track interruptions divided by the estimated number of underlying
trajectories (inferred from track lifetimes, not from labels). The proxy rewards configurations that
produce many long, unbroken tracks and penalizes fragmentation; it requires no
target-domain labels. We optimize \eqref{eq:bo-proxy} for 50 trials per sequence,
reporting the HOTA of the proxy-maximizing configuration. 

\section{Results}
\label{sec:results}

Here we address four questions: how brittle is ground-truth-supervised
hyperparameter transfer (Section~\ref{sec:results-brittleness}); does
VLM-guided adaptation recover the lost performance
(Section~\ref{sec:results-recovery}); which agent and which signal modality
drive the gain (Section~\ref{sec:results-ablation}); and where does the
approach fail (Section~\ref{sec:results-failure})?


\begin{table}[t]            
  \centering
  \label{tab:bo_comparison}
  \begin{tabular}{llcccc}
  \toprule
  \textbf{Oracle Source} & \textbf{Target} & \textbf{Target Oracle} & \textbf{Oracle xfer} & \textbf{BO-proxy} & \textbf{VLM (ours)} \\
  \midrule
    MOT17 & MOT20 & 0.357 & 0.267 & 0.305 & \textbf{0.328} \\
    MOT20 & MOT20 & 0.357 & 0.338 & 0.322 & \textbf{0.349} \\
    DT & MOT20 & 0.357 & 0.312 & 0.305 & \textbf{0.326} \\

    MOT17 & DT & 0.501 & \textbf{0.467} & 0.431 & 0.452 \\
    MOT20 & DT & 0.501 & 0.428 & 0.412 & \textbf{0.441} \\
    DT & DT & 0.501 & 0.460 & 0.402 & \textbf{0.477} \\
    
    MOT17 & MOT17 & 0.457 & \textbf{0.440} & 0.285 & 0.433 \\
    MOT20 & MOT17 & 0.457 & \textbf{0.430} & 0.399 & 0.428 \\
    DT & MOT17 & 0.457 & \textbf{0.438} & 0.322 & 0.430 \\

  \bottomrule
  \end{tabular}
    \caption{Cross-domain transfer and recovery. We compute a source domain configuration oracle for each of MOT17, MOT20 and DanceTrack (DT). Each row shows four metrics for each combination of source→target domain transfer: The oracle computed on the target domain (near-optimal ceiling for that domain); the results of transferring the source domain oracle to that new domain; the results of a Bayesian Optimization of the new target domain using the proxy metric in Equation~\ref{eq:bo-proxy}; and finally the results of allowing the VLM to optimize tracker performance on the new domain, starting from the transferred source oracle. Our VLM-based tracker tuner excels at overcoming large domain shifts, such as between MOT17→MOT20 and MOT17/20→DanceTrack, while avoiding catastrophic regressions for transfers where the starting conditions are already near optimal such as all →MOT17 transfers. For same-domain rows (e.g.\ MOT17$\rightarrow$MOT17), \emph{oracle xfer} applies
a single configuration pooled across the source sequences, whereas \emph{target
oracle} averages per-sequence fits; the diagonal gap between them therefore
reflects per-sequence specialization, not domain shift.}
\label{tab:cross-domain}
\end{table}

\begin{table*}[t]
  \centering
  \begin{tabular}{@{}llccccc@{}}
    \toprule
    Source & Condition & 01 & 02 & 03 & 05 & \textbf{Mean} \\
    \midrule
    MOT17      & Oracle transfer         & 0.427 & 0.356 & 0.190 & 0.095 & 0.267 \\
    MOT20      & Oracle transfer         & \textbf{0.446} & 0.388 & 0.312 & 0.206 & 0.338 \\
    DanceTrack & Oracle transfer         & 0.424 & 0.374 & 0.289 & 0.160 & 0.312 \\
    \midrule
    MOT17      & VLM from oracle  & 0.429 & 0.360 & 0.327          & 0.195          & 0.328 \\
    MOT20      & VLM from oracle  & \textbf{0.446} & \textbf{0.393} & 0.312          & \textbf{0.247} & \textbf{0.349} \\
    DanceTrack & VLM from oracle  & 0.407 & 0.369 & \textbf{0.333} & 0.195          & 0.326 \\
    \midrule
    \multicolumn{2}{@{}l}{\textit{Target per-sequence oracle (GT ceiling)}} & 0.459 & 0.372 & 0.348 & 0.248 & 0.357 \\
    \bottomrule
  \end{tabular}
  \caption{MOT20 domain transfer per sequence (HOTA). The first three rows show the performance of various source oracles transferred to each of the MOT20 video sequences, which HOTA numbers displayed across the columns. The bottom 3 rows show the results of running our VLM tracker tuner from these oracle transfer starting points, recovering performance significantly on the more complex scenes (MOT20-03 and -05). Bold
marks the best non-ceiling result per column.}
  \label{tab:mot20-per-sequence}
\end{table*}

\subsection{Brittleness of Oracle Transfer}
\label{sec:results-brittleness}

Table~\ref{tab:cross-domain} reports mean HOTA for all $3\times3$
source/target pairs. Cross-domain oracle transfer underperforms the target
per-sequence oracle ceiling in every row, and the shortfall grows with the
magnitude of the domain shift: denser crowds (MOT17$\rightarrow$MOT20) and
faster motion (MOT17/MOT20$\rightarrow$DanceTrack) both widen the gap. The
most severe case is MOT17$\rightarrow$MOT20, a $0.090$ HOTA shortfall
($0.267$ vs.\ a $0.357$ ceiling).

This mean shortfall hides extreme per-sequence variation
(Table~\ref{tab:mot20-per-sequence}). The moderate-density sequences MOT20-01
and MOT20-02 lose under $0.032$ HOTA, while the dense MOT20-03 and MOT20-05
fall catastrophically, more than $0.15$ HOTA below their target oracles, and
together account for the majority of the cross-domain degradation. The
mechanism is detection-stage under-coverage: MOT17's per-sequence oracle
settles on \texttt{conf\_thresh} in the $0.20$--$0.30$ range, well-tuned for
MOT17's moderate density but high enough to reject a large fraction of valid
detections in dense MOT20 crowds.

\begin{table*}[t]
  \centering
\begin{tabular}{lccccc}
\toprule
M17$\rightarrow$M20 & 01 & 02 & 03 & 05 & Mean \\
\midrule
Oracle xfer & .427 & .356 & .190 & .095 & .267 \\
VLM (m$\pm\sigma$) & .406{\tiny$\pm$.006} & .346{\tiny$\pm$.006} & .320{\tiny$\pm$.017} & .192{\tiny$\pm$.002} & .316{\tiny$\pm$.005} \\
\bottomrule
\end{tabular}
\caption{We re-run the M17$\rightarrow$M20 5 times to determine whether the performance gains at temperature 1 are noise. The largest gains, in MOT20-03 and -05, clear $2\sigma$ while the simpler videos -01 and -02 a stable tendency not to drift parameters away from good starting points.}
\end{table*}
\subsection{VLM-Guided Recovery}
\label{sec:results-recovery}

For each source/target pair we initialize the tracker with the source-domain
oracle and run the two-phase adaptation loop (Section~\ref{sec:loop}) on the
target sequences, using no target-domain labels. On MOT17$\rightarrow$MOT20,
adaptation lifts mean HOTA from $0.267$ to $0.328$ ($+0.061$), recovering
$67.8\%$ of the $0.090$ shortfall and landing within $0.029$ of the ceiling.
The gains concentrate exactly where the brittleness did
(Table~\ref{tab:mot20-per-sequence}): MOT20-03 and MOT20-05 recover $86.7\%$
and $65.4\%$ of their per-sequence headroom, while the already-near-ceiling
moderate sequences move little. This is the expected behavior of a system
whose diagnostic signal scales with the visibility of failure modes: severe
mis-detection gives the VLM dense visual evidence to act on; a near-correct
configuration gives it little.

This produces a useful asymmetry across the matrix
(Table~\ref{tab:cross-domain}). On high-shift transfers
(MOT17$\rightarrow$MOT20, MOT20$\rightarrow$DanceTrack,
DanceTrack$\rightarrow$MOT20) the VLM closes most of the gap; on low-shift
transfers ($\rightarrow$MOT17), where oracle transfer is already within
$\sim$0.010 of the ceiling, it produces only small changes
($|\Delta|<0.010$). The single substantial regression,
MOT17$\rightarrow$DanceTrack ($-0.015$), is analyzed in
Section~\ref{sec:results-failure}.

The Bayesian-optimization baseline (\emph{BO-proxy}), also label-free,
optimizes the same parameter space against the scalar proxy of
Equation~\ref{eq:bo-proxy}. It recovers part of the high-shift gap but is
outperformed by the VLM on every $\rightarrow$MOT20 and
$\rightarrow$DanceTrack sequence. The revealing contrast is on low-shift
transfers: BO-proxy regresses sharply (e.g.\ $0.440\rightarrow0.285$ on
MOT17$\rightarrow$MOT17), while the VLM holds steady. A black-box optimizer
has no notion of ``leave the configuration alone''. It searches until the
proxy is satisfied, and a miscalibrated proxy drives it off a good operating
point. The VLM instead returns \texttt{clean}/\texttt{no\_lever} when its
diagnostics show no failure mode, which is what produces the no-harm
behavior.

\begin{table*}[t]
  \centering
  \begin{tabular}{@{}lccccc r@{}}
    \toprule
    Condition & 01 & 02 & 03 & 05 & \textbf{Mean}\\
    \midrule
    Transfer from MOT17 Oracle & 0.427 & 0.356 & 0.190 & 0.095 & 0.267 \\

    \quad + Detection agent only        & 0.398 & 0.335 & 0.312 & 0.190 & 0.309\\
    \quad + Tracking agent only         & 0.402 & 0.346 & 0.230 & 0.131 & 0.277\\
    \quad + Both agents & \textbf{0.429} & \textbf{0.360} & \textbf{0.327} & \textbf{0.195} & \textbf{0.328}\\
    \midrule
    \textit{Target per-sequence oracle} & 0.459 & 0.372 & 0.348 & 0.248 & 0.357 & \\
    \bottomrule
  \end{tabular}
  \caption{Agent Ablation: MOT17${\to}$MOT20 transfer.
  All VLM conditions start from the MOT17 oracle config (same as oracle transfer).
  \emph{Detection agent only} tunes $\{\mathrm{conf\_thresh}, \mathrm{nms\_thresh}\}$;
  \emph{tracking agent only} tunes $\{\mathrm{match\_thresh}, \mathrm{spawn\_thresh},
  \mathrm{max\_lost\_age}, \mathrm{velocity\_decay}, \mathrm{min\_hits}, \mathrm{appearance\_thresh} \}$.
  The detection agent is responsible for the majority of the gain ($+$0.042 of $+$0.061 total), concentrated on the two hardest sequences (MOT20-03/05) where conf mismatch is catastrophic. The tracking agent contributes $+$0.010 in isolation and a marginal $+$0.019 when applied after detection, and the agents compose near-additively ($0.042 + 0.010 \approx 0.061$). Bold marks the best non-ceiling result per column.}
  \label{tab:mot20-ablation}
\end{table*}

\begin{table}[t]
\centering
\small
\begin{tabular}{lccccc}
\toprule
Condition & 01 & 02 & 03 & 05 & Mean \\
\midrule
Oracle transfer & 0.427 & 0.356 & 0.190 & 0.095 & 0.267 \\
VLM text-only & 0.407 & 0.348 & 0.235 & 0.131 & 0.281 \\
VLM visual-only & \textbf{0.449} & 0.340 & 0.312 & \textbf{0.219} & \textbf{0.330} \\
VLM text+vision & 0.429 & \textbf{0.360} & \textbf{0.327} & 0.195 & 0.328 \\
\midrule
Target oracle (ceiling) & 0.459 & 0.372 & 0.348 & 0.248 & 0.357 \\
\bottomrule
\end{tabular}
\caption{Signal modality ablation on MOT17$\rightarrow$MOT20.
Text-only (scalars without visual panels) barely improves on oracle
transfer; visual-only (panels without scalars) matches the full
system within noise. The text-only agent improves the hardest
sequences modestly via tracking-side intervention but cannot
diagnose the detection-stage failures that drive the cross-domain
gap. Bold marks the best non-ceiling result per column.}
\label{tab:modality_ablation}
\end{table}

\subsection{Agent Ablation: Where Do the Gains Come From?}
\label{sec:results-ablation}

To attribute the MOT17$\rightarrow$MOT20 recovery across the two phases, we
run each agent in isolation from the same MOT17 oracle initialization
(Table~\ref{tab:mot20-ablation}). The detection agent alone recovers $+0.042$
HOTA, the majority of the $+0.061$ total, concentrated on the two hardest
sequences. The tracking agent alone recovers only $+0.010$: the MOT17
oracle's tracking parameters (\texttt{match\_thresh}=0.90,
\texttt{max\_lost\_age}=15) are not catastrophically mistuned for MOT20's
pedestrian-speed motion, so it has less to correct. 
Run in sequence, the two phases compose near-additively: the tracking agent's
marginal contribution rises from $+0.010$ in isolation to $+0.019$ when
applied after detection ($0.042 + 0.019 \approx 0.061$), a small positive
interaction. The tracking phase does not undo the detection phase's
gains, validating the sequential ordering. On moderate-density sequences (MOT20-01: $-0.029$, MOT20-02: $-0.021$) the
detection agent slightly over-tunes, which is the same noise-floor behavior examined
in Section~\ref{sec:results-failure}. On
MOT17$\rightarrow$MOT20, then, adaptation is fundamentally a detection-phase
intervention.

We further separate the contribution of visual evidence from scalar metrics
with two restricted variants: \textit{visual-only} omits the scalar
calibration metrics, \textit{text-only} omits the visual panels
(Table~\ref{tab:modality_ablation}). Visual-only matches the full system
($0.330$ vs.\ $0.328$); text-only reaches only $0.281$, barely above oracle
transfer. The traces explain why: scalar metrics like
\texttt{near\_threshold\_fraction} cannot distinguish ``operating point is
fine'' from ``operating point catastrophically misses pedestrians in a dense
crowd'' as the same numeric signature arises from either regime. The
text-only agent therefore barely touches detection parameters and acts mostly
on tracking, which helps modestly but cannot recover the detection-stage
failure driving the gap. The operative signal in our framework is the
structured visual evidence, not the scalars. A classical optimizer
over those same scalars, like BO-proxy and the text-only agent, would be
equally blind to the detection failures that dominate cross-domain
brittleness.


\begin{table}[t]
  \centering
  \begin{tabular}{@{}lcccccccc c r@{}}
    \toprule
    Condition & 04 & 05 & 07 & 10 & 19 & 34 & 47 & 58 & \textbf{Mean} \\
    \midrule

    MOT17 Oracle 
      & \textbf{0.51} & 0.48 & \textbf{0.51} & \textbf{0.61} & 0.32 & 0.38 & 0.33 & \textbf{0.59} & \textbf{0.47}  \\

    \quad + Det agent 
      & 0.34 & \textbf{0.56} & 0.50 & 0.47 & 0.27 & 0.38 & 0.30 & 0.55 & 0.42 \\
    \quad + Tracking agent 
      & 0.39 & \textbf{0.56} & 0.49 & 0.47 & 0.28 & 0.36 & 0.37 & 0.44 & 0.42 \\
    \quad + Both agents 
      & 0.35 & 0.43 & 0.50 & \textbf{0.61} & \textbf{0.34} & \textbf{0.39} & \textbf{0.40} & \textbf{0.59} & 0.45\\
    \bottomrule



  \end{tabular}
  \caption{Agent Ablation: MOT17${\to}$DanceTrack transfer (HOTA).
  All VLM conditions start from the MOT17 oracle config.
  Unlike the MOT20 case, \emph{neither agent alone improves} on oracle transfer ($-$0.046 each):
  the MOT17 oracle (match$=$0.90, age$=$15) is already near-optimal for DanceTrack's fast motion,
  so each isolated agent perturbs away from it without sufficient feedback to self-correct.
  The full two-phase loop limits the regression to $-$0.015 through inter-phase correction.
  DT-0047 is the revealing exception: the tracking agent alone yields $+$0.033 over oracle transfer
  (0.367 vs.\ 0.334) while the detection agent hurts ($-$0.033).
  This sequence has reliable detections but identity fragmentation. The tracking agent correctly
  tightens match/spawn, but the detection agent incorrectly intervenes on conf and destroys the gain.
  Bold marks the best VLM result per column.}
  \label{tab:ablation_dancetrack}
\end{table}

\subsection{Failure Case: MOT17$\rightarrow$DanceTrack}
\label{sec:results-failure}

MOT17$\rightarrow$DanceTrack is the only matrix cell where adaptation
regresses substantially. Both isolated agents regress by an identical
$-0.046$ HOTA, and the full loop limits the regression to $-0.015$ through
inter-phase correction (Table~\ref{tab:ablation_dancetrack}), the mirror
image of the MOT20 case, where both agents helped and the loop accumulated
the gains. The cause is that the MOT17 oracle is well-matched to the target:
its tracking parameters suit DanceTrack's fast motion, and detection coverage
is adequate at the source threshold since scene density is comparable. With
no clear failure mode to diagnose, the agents nonetheless act on local visual
residuals and perturb an already near-optimal configuration.

DT-0047 illustrates the mechanism: detection is reliable but identity
fragmentation dominates. The tracking agent correctly tightens
\texttt{match\_thresh}/\texttt{spawn\_thresh} ($+0.033$), while the detection
agent, lacking a detection-stage failure to act on, intervenes on
conf\_thresh and loses $-0.033$; the full loop's tracking phase
partially repairs the detection-phase damage ($+0.064$ overall). It is the
clearest case where two-phase composition matters. A single agent helps or
hurts decisively depending on which failure mode is present.

This is a failure of applicability, not of mechanics: the agents return
well-formed diagnoses and the meta-agent reconciles them correctly; the
system is simply being asked to improve a configuration already near a local
optimum. Synthesizing across Sections~\ref{sec:results-recovery}--%
\ref{sec:results-failure}, VLM-guided tuning is effective when two conditions
hold: the domain shift produces visible failure modes that map onto the
exposed parameters, and the source configuration is mistuned enough that the
adaptation signal exceeds the noise floor of agent intervention.
MOT17$\rightarrow$MOT20 satisfies both; MOT17$\rightarrow$DanceTrack
satisfies neither. This suggests a deployment heuristic: apply VLM tuning when
the source-target gap is large and detection-concentrated, and skip it when
the source transfers well, which a practitioner can estimate from a
small held-out set or even qualitative inspection before committing to the
adaptation.

\section{Conclusion}
\label{sec:conclusion}

We presented a system for cross-domain tracker
adaptation that uses a Vision-Language Model as a
diagnostic agent operating over rendered tracker
output, calibration metrics, and a structured
decision procedure encoded in the agent's system
prompt. The system requires no target-domain labels
to tune the exposed parameter space of a
fixed detect-to-track pipeline.

Our results establish three claims. First, source-domain oracle configurations
under-perform target-domain oracles by up to
$0.158$ HOTA on individual sequences  for the
MOT17 to MOT20 transfer, with the
degradation concentrating on sequences whose
density regime differs from the source. Second,
VLM-guided adaptation from a source-oracle
initialization recovers $67.8\%$ of the
per-sequence oracle headroom on
MOT17$\rightarrow$MOT20 and up to $86.7\%$ on the
highest-density target sequence, without target
labels. 
Third, the recovery exhibits an asymmetric profile across the transfer matrix.
We achieve substantial gains when the source configuration transfers poorly, yet no
significant perturbation when it transfers well. Whereas, a label-free
Bayesian-optimization baseline driven by a scalar proxy can degrade
near-optimal configurations (e.g.\ $0.440 \rightarrow 0.285$ on
MOT17$\rightarrow$MOT17). The single substantive regression
(MOT17$\rightarrow$DanceTrack) is due to a source configuration that
is already near-optimal for the target.

The method has limits that the results make
explicit. VLM-guided adaptation only acts within
the exposed parameter space of the underlying
tracker; domain shifts that demand changes to
detector weights, association features, or
appearance models are outside its reach. The system
also requires source-domain labels to compute the
oracle initialization, so the contribution is best
understood as label-free \emph{target} adaptation
rather than label-free tuning end-to-end. Finally,
the agents can perturb already-good configurations
when no clear failure mode is present, producing
small regressions on low-shift transfers; the
two-phase composition mitigates but does not
eliminate this behavior.

Several directions merit further study. First, within the domain of tracking system, we only explore here a small subset of the tools, parameters, models and other tuning variables that could be exposed to the VLM. Secondly, there is a large exploration space remaining of methods fo visualizing tracker performance for the tuning model. Lastly, we encourage the community to explore additional perception tasks which suffer from similar open-loop failure modes, and which could benefit from automated VLM tuning. 


\bibliography{egbib}
\end{document}